\documentclass[journal,twoside,web]{ieeecolor}
\usepackage{tmi}
\usepackage{cite}
\usepackage{amsmath,amssymb,amsfonts}

\usepackage{algorithm,ulem}
\usepackage{algpseudocode}%
\usepackage{fancyhdr}
\usepackage[algo2e]{algorithm2e}

\newtheorem{theorem}{Theorem}
\newtheorem{lemma}[theorem]{Lemma}

\usepackage{graphicx}
\usepackage{textcomp}
\usepackage{bbm}
\usepackage{bm}

\usepackage{multirow}  
\usepackage{booktabs} 
\usepackage{threeparttable}
\def\BibTeX{{\rm B\kern-.05em{\sc i\kern-.025em b}\kern-.08em
    T\kern-.1667em\lower.7ex\hbox{E}\kern-.125emX}}
\usepackage[table]{xcolor}

\usepackage[misc]{ifsym}
\usepackage{gensymb}
\usepackage{url}

\newcommand{\etal}{\emph{et al.}}

\definecolor{FigureColor}{RGB}{0,138,218}
\newcommand{\fig}[1]{\textcolor{FigureColor}{Fig.#1}}

\definecolor{EqColor}{RGB}{0,138,218}
\newcommand{\eqn}[1]{\textcolor{EqColor}{Eq.{#1}}}

\definecolor{TabColor}{RGB}{0,138,218}
\newcommand{\tab}[1]{\textcolor{TabColor}{Table {#1}}}

\definecolor{Equation}{RGB}{0,138,218}
\definecolor{Table}{RGB}{0,138,218}
\definecolor{Algorithm}{RGB}{0,138,218}
\definecolor{Annotation}{RGB}{19, 139, 44}
\begin{document}
\title{Differentiable Topology-Preserved Distance Transform for Pulmonary Airway Segmentation}
\author{Minghui Zhang, Guang-Zhong Yang, \IEEEmembership{Fellow, IEEE}, Yun Gu, \IEEEmembership{Member, IEEE}
\thanks{The manuscript is received at 2022. This work was partly supported by National Key R\&D Program of China (2019YFB1311503), Shanghai Sailing Program (20YF1420800), NSFC (62003208).}
\thanks{Minghui Zhang (minghuizhang@sjtu.edu.cn), Yun Gu (geron762@sjtu.edu.cn) and Guang-Zhong Yang (gzyang@sjtu.edu.cn) are 
with Institute and Medical Robotics, Shanghai Jiao Tong University, 200240, Shanghai, CHINA.}%
}

\maketitle

\begin{abstract}
Detailed pulmonary airway segmentation is a clinically important task for endobronchial intervention and treatment of peripheral 
located lung cancer lesions. Convolutional Neural Networks (CNNs) are promising tools for medical image analysis but have been performing 
poorly for cases when existing a significant imbalanced feature distribution, which is true for the airway data as the trachea and principal 
bronchi dominate most of the voxels whereas the lobar bronchi and distal segmental bronchi occupy a small proportion. 
In this paper, we propose a Differentiable Topology-Preserved Distance Transform 
(DTPDT) framework to improve the performance of airway segmentation. A Topology-Preserved 
Surrogate (TPS) learning strategy is first proposed to balance the training progress within-class distribution. 
Furthermore, a Convolutional Distance Transform (CDT) is designed to identify the breakage phenomenon with 
superior sensitivity and minimize the variation of the distance map between the prediction and ground-truth. 
The proposed method is validated with the publically available reference airway segmentation datasets.
The detected rate of branch and length on public EXACT'09 and BAS datasets are 82.1\%/79.6\% and 96.5\%/91.5\% respectively, 
demonstrating the reliability and efficiency of the method in terms of improving the topology completeness 
of the segmentation performance while maintaining the overall topology accuracy. 
\end{abstract}


\section{Introduction}
\label{sec:introduction}
\IEEEPARstart{A}{irway} segmentation is a crucial foundation for the diagnosis, and treatment of pulmonary diseases including asthma, bronchiectasis, and emphysema. Accurate segmentation based on computed tomography (CT) enables quantitative measurements of airway dimensions and wall thickness. For treatment, the extraction of the airway model from CT images is a prerequisite for both pre- and intra-operative navigation in endobronchial interventions.

\begin{figure*}[!t]
\centering{\includegraphics[width=1.0\linewidth]{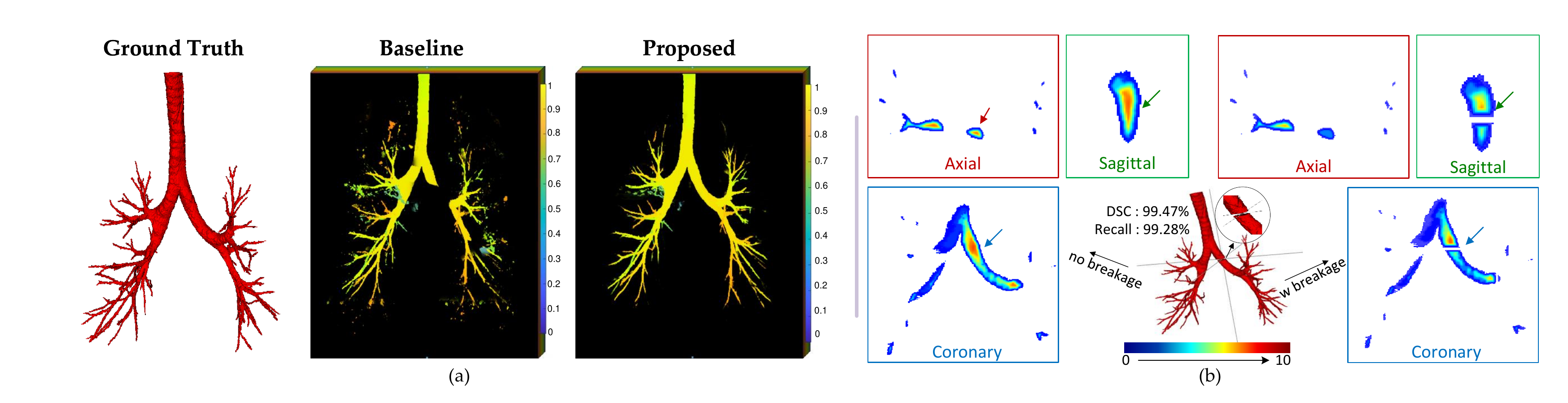}}
\caption{Illustrations of challenges in airway segmentation: (a) Maximum intensity projection of the probability maps obtained from different methods. (b) Comparison between the distance transform map of the binary result with or without the breakage. In (a), the baseline generates a skew probability map that peripheral airways share lower confidence probability than that of the principal bronchi, while this problem is alleviated by the proposed method. It can be observed from (b) that the small breakage only causes slight effects on the DSC/Recall metric while revealing significant differences in the distance transform map.}
\label{fig::dist_trans}
\end{figure*}

Due to the fine-grained pulmonary airway structure, manual annotation is time-consuming, error-prone, and highly relies on the expertise of clinicians. To alleviate such burdens and expedite the explorations of airways, automatic airway segmentation algorithms are being continuously pursued. 
Thus far, public airway datasets with annotation (EXACT'09\cite{EXACT09}, BAS\cite{qin2019airwaynet}) have been widely used to develop automatic 
airway extraction algorithms~\cite{Bruijne_Juarez_Automatic,Bruijne_3DUnet-Graph,jin20173d,wang2019tubular,qin2019airwaynet,zheng2021alleviating}. 
The EXACT'09 challenge has established standard evaluation metrics for airway segmentation performance. 
These metrics are mainly categorized into two groups including overlap based accuracy and topological measurements. 
For overlap based accuracy measure, the Dice Similarity Coefficient (DSC) measure is used to represent the segmentation accuracy. 
Topological measurements include the tree length detected rate (TD, \%) and branch detected rate (BD, \%), which are used to 
measure the topological completeness and continuity of the predictions. 
Unlike the general segmentation tasks which focus more on overlap based accuracy, the topological measurements are also important 
for airway segmentation as it reflects the intrinsic structure of the airway tree, 
which is particularly relevant for intra-operative navigation under physiological motion including respiration.
In practice, ensuring topological accuracy is challenging as it is difficult to be achieved by using only the overlap based loss functions, 
e.g., Dice loss. This can be attributed to the severe within-class distribution imbalance, 
as the trachea and principal bronchi dominate most of the voxels while the lobar bronchi and distal segmental bronchi only 
occupy a small proportion. It is known that CNNs trained with class-imbalanced data may perform poorly 
on the minor classes with scarce training data~\cite{buda2018systematic}. Consequently, as shown in \fig{\ref{fig::dist_trans}(a)} as an example, 
the baseline method trained with dice loss function generates a skew likelihood map that peripheral airways share a lower confidence probability than 
that of the principal bronchi. Although variants of dice functions, e.g. clDice~\cite{shit2021cldice} and GUL~\cite{zheng2021alleviating}, 
can increase the sensitivity of the minor classes, they still face the over-segment problems. 

In practice, even if the recall is higher, the overlap based loss functions are imperceptive to small breakages of the airway structures. 
To illustrate this effect, we present the distance transformation map of the prediction in \fig{\ref{fig::dist_trans}(b)} 
where voxel intensities show the distance to the nearest boundary. It can be observed that the breakage merely induces marginal voxel-level 
errors while the topology connectivity is changed. Hence, single supervision from the overlap based loss functions cannot 
guarantee a high topological accuracy.

To resolve the aforementioned issues, we propose in this paper a Differentiable Topology-Preserved Distance Transform (DTPDT) framework to facilitate 
the performance of airway segmentation. \uline{A Topology-Preserved Surrogate (TPS) learning strategy is first proposed to balance the training progress within-class distribution of airways.} As observed by Qin \etal~\cite{qin2021learning}, the prediction binarized by a smaller threshold can boost a significant improvement of the TD/BD, while the DSC drops dramatically. As the threshold is relaxed, the recall metric of the foreground class will increase, and therefore the breakage problem is alleviated and the TD/BD is increased. However, the oversimplified relaxation of the threshold inevitably degrades the precision and the DSC metric, which is detrimental to the quantitative measurement analysis. This phenomenon is equivalent to the gradient erosion and dilation problem introduced by Zheng \etal~\cite{zheng2021alleviating}. Instead of using the trivial multi-stages training procedure in previous works~\cite{zheng2021alleviating,yu2022break}, the proposed Topology-Preserved Surrogate (TPS) aims to balance the trade-off between the topology completeness and topology correctness via independent objective functions. The topology completeness objective function is inspired by previous work~\cite{zheng2021alleviating,wang2019tubular}, we designed a compound loss function embedded with distance prior to adjust the importance of each voxel, reinforcing the network to detect more airway branches. Meanwhile, the topology correctness objective function deals with the underlying over-segment problem via maximizing the area under the Precision-Recall curve.

\uline{Futhermore, the Convolutional Distance Transform (CDT) is proposed to pay attention to the broken connection.} The breakage merely induces marginal voxel-level errors but causes the catastrophic topology mistakes. As demonstrated in \fig{\ref{fig::dist_trans}(b)}, the distance transform can highlight breakage and enforce the segmentation result to have the same distance map as the ground-truth. Unlike previous works~\cite{bui2019multi,navarro2019shape} that need to learn the distance transform, we approximate the Euclidean distance from the probability prediction and construct the distance map loss function. The CNNs trained with such loss can achieve satisfactory topological fidelity without sacrificing the voxel-wise accuracy. 

The proposed DTPDT is an end-to-end framework that does not require a multi-stage training procedure. The two critical components, TPS and CDT cooperate with each other, aiming to facilitate the performance of airway segmentation under multiple evaluation criteria. Extensive experiments on two public pulmonary airway datasets demonstrate that the proposed method has achieved superior performance compared to other state-of-the-art approaches.

\section{Related Work}
\subsection{Airway Segmentation}
To relieve the burden of manual delineation and help clinicians explore the influence of pneumonia on airways, automatic pulmonary airway segmentation algorithms have been widely explored over the decades. In 2009, EXACT'09~\cite{EXACT09} challenge provided a platform for comparing airway extraction algorithms using a public dataset and standard evaluation metrics. During that period, several methods using multi-thresholds~\cite{van2009automatic}, template matching~\cite{born2009three}, and region growing~\cite{lo2010vessel} have been proposed to automatically segment the airways. However, these methods often fail in extracting the smaller peripheral bronchi due to the lack of discriminative features.

Recently, the progress of deep learning has promoted the research on airway segmentation. Juarez \etal~\cite{Bruijne_Juarez_Automatic} directly adopted 3D CNNs with an elaborated pipeline for automatic airway segmentation. 3D UNet cooperating with the graph refinement~\cite{jin20173d}, attention mechanism~\cite{qin2020learning}, and tubular structural distance loss~\cite{wang2019tubular} was also proposed to extract more discriminative features. The connectivity of the airway prediction also raises attention. Qin \etal~\cite{qin2019airwaynet} proposed the AirwayNet that transformed the binary segmentation task into 26-neighborhood connectivity prediction. Wu \etal\cite{wu2022ltsp} utilized the long-range slice continuity information to preserve the topology completeness. Both Zheng \etal~\cite{zheng2021alleviating} and Yu \etal~\cite{yu2022break} 
adopted the WingsNet~\cite{zheng2021alleviating} as backbone with multi-stage training procedure. A general union loss (GUL) was further designed by Zheng \etal~\cite{zheng2021alleviating} to alleviate the within-class distribution imbalance. Yu \etal~\cite{yu2022break} resolved the problem via a breakage-sensitive loss function. However, the topology-preserving problem of the airway has not been discussed thoroughly.

\subsection{Distance Transform}
Given an image $I$ and the corresponding ground-truth of segmentation $G$, the distance transform ($G_{\mathrm{DT}}$) is defined as:
\begin{align}
G_{\mathrm{DT}}=\left\{\begin{array}{ll}
\inf\limits_{z \in \partial G}\left\|x-z\right\|_2,& x \in G_{\mathrm{in}},\\
0,& \mathrm{others,}\label{eq::Distance_Transform}
\end{array}\right.
\end{align}
where $\left\|x-z\right\|_2$ denotes Euclidian distance between voxels $x$ and $z$. $\partial G$ and $ G_{\mathrm{in}}$ represent the boundary and inside of the ground-truth, respectively. The distance transform can highlight the local structure of segmentation masks, thus providing alternative supervision to the CNNs. Recently, some works are dedicated to designing new loss functions based on the distance transform map. For example, Kervadec \etal \cite{kervadec2019boundary} designed the boundary loss to calculate boundary variations between prediction result and ground-truth via an integral approach rather than the complex local differential computations. Xue \etal \cite{xue2020shape} directly regressed the signed distance function (SDF) and proposed the corresponding loss to penalize the output SDF with the wrong sign. Other works designed multi-task settings based on the distance transform map and the ground-truth, which add auxiliary tasks to augment CNNs. The reasonable explanation for the advantage of the distance transform is that it can introduce the shape prior knowledge to the CNNs. However, as pointed out by Wang \etal \cite{wang2020deep}, the direct regression of the distance transform is unstable. Hence, they rephrased the distance map prediction as a classification task based on quantization. Although their framework can predict the segmentation and the distance map simultaneously, a geometry-aware refinement procedure is still needed. In our work, we proposed the differentiable distance transform that functions at the likelihood map and then constructed a distance map loss function for topology-preserving usage.

\begin{figure*}[t]
\centering{\includegraphics[width=0.9\linewidth]{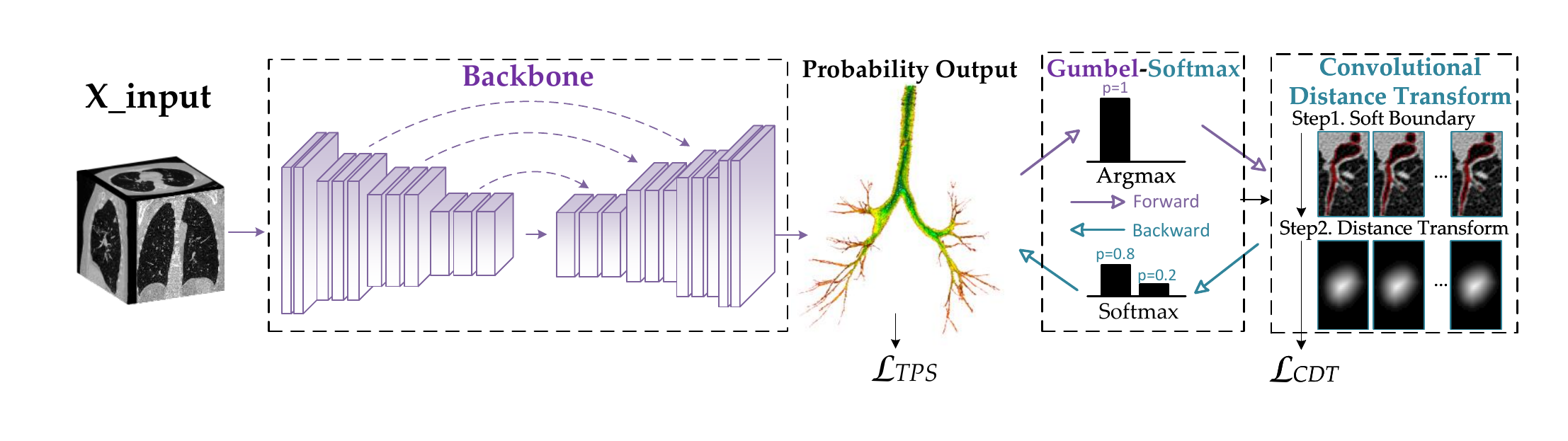}}
\caption{Framework of the proposed Differentiable Topology-Preserved Distance
Transform (DTPDT) network. The backbone (3D UNet) generates the probability output, followed by two supervised branches that are used to guide the optimization schedule. The first branch is supervised by the Topology-Preserved Surrogate (TPS) which is designed to achieve satisfactory trade-off between topology completeness and topology correctness via independent objective functions. The second branch is the proposed Convolutional Distance Transform (CDT) which approximates the Euclidean distance from
the probability output via differentiable operations. The objective function of CDT is constructed to achieve high topological fidelity without sacrificing the voxel-wise
accuracy.}
\label{MainFramework}
\end{figure*}

\section{Methods}
\subsection{Problem Formulation and Method Overview}
This work addressed the airway segmentation problem. $\mathbf{x}\in\mathbf{X}$ is the input volume and  $\mathbf{y}\in\mathbf{Y}$ is the ground-truth of segmentation mask. We design a model $\mathbf{\hat{y}} = \mathcal{F}(\mathbf{x}, \Theta)$ where $\Theta$ represents the model parameters and $\mathbf{\hat{y}}$ is the likelihood map of prediction. The model is optimized by loss function $\mathcal{L(\mathbf{\hat{y}}, \mathbf{{y}})}$ which measures the difference between the ground-truth and predictions. \fig{\ref{MainFramework}} illustrates the basic framework of this work. The 3D UNet~\cite{cciccek20163d} is deployed as the backbone followed by two critical components, TPS and CDT. During training procedure, TPS and CDT function upon the probability output of the 3D UNet, cooperating to preserve the topology completeness and correctness of airway. 


\subsection{Topology-Preserved Surrogate}
Commonly evaluated metrics for pulmonary airway segmentation\cite{EXACT09}, such as tree length detected rate (TD) and branch detected rate (BD), are non-differentiable that 
hard to be directly optimized through stochastic gradient descent methods. With the advance of the surrogate loss function~\cite{LearnSurrogates, IOULoss}, we propose a Topology-Preserved Surrogate (TPS) module, aiming to achieve both high topology completeness and topology correctness. 

\begin{figure}[!t]
\centering{\includegraphics[width=1.0\linewidth]{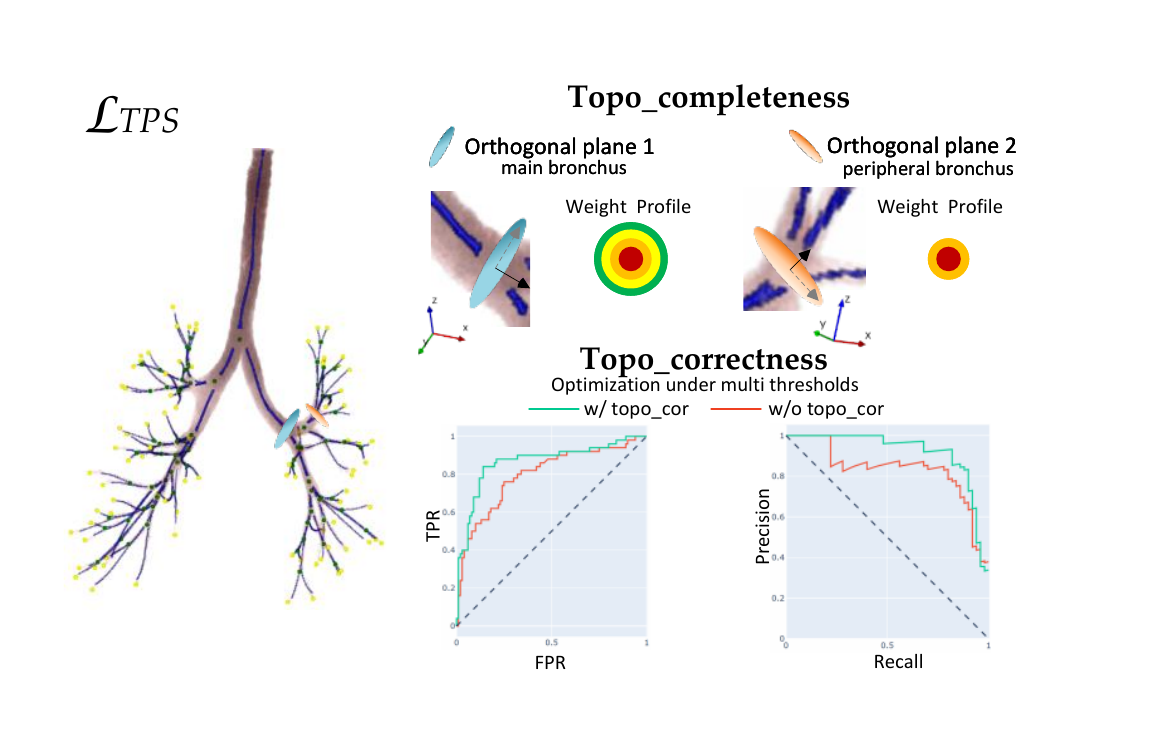}}
\caption{The TPS module preserves both topology completeness and correctness. For topology completeness, we set different weight profiles by considering the radius of airway branches. For topology correctness, we maximize the area under precision-recall curve to improve the accuracy with multiple thresholds.}
\label{fig::tps}
\end{figure}

\textbf{Topology Completeness}: The difficulty of improving the topology completeness of the pulmonary airway segmentation can be attributed to within-class distribution imbalance\cite{zheng2021alleviating}. Large airways occupy the majority of foreground voxels, and such imbalanced distribution affects the data-driven deep learning methods, 
which may lead to poor performance on the peripheral bronchi. The Tversky loss~\cite{salehi2017tversky} function, $ \mathcal{L}_{\tiny{tversky}}$,  is first deployed to enhance the sensitivity for peripheral airways:
\begin{align}
    \begin{split}
    \mathcal{L}_{\tiny{tversky}} = 1 - \frac{\sum_{i = 1}^{N} \mathbf{\hat{y}}_{i}\mathbf{y}_{i}}
    {\alpha_{t}\sum_{i}\mathbf{\hat{y}}_{i} + \beta_{t}\sum_{i}\mathbf{y}_{i}}, \label{Loss_Tversky}
    \end{split}
\end{align}
where $\alpha_{t} + \beta_{t} = 1$. Both $\alpha_{t}$ and $\beta_{t}$ are hyper-parameters to balance the recall and sensitivity of segmentation.  However, this can lead to serve dilation problem~\cite{zheng2021alleviating} with a constant $\alpha_{t}$. As pointed out by Zheng \etal\cite{zheng2021alleviating}, the gradients of different airway voxels should vary with the branch sizes during backward propagation. Therefore, the tubular radius prior is integrated into the Cross-Entropy (CE) loss function to further resolve the within-class distribution imbalance as follows:
\begin{align}
    \begin{split}
    \mathcal{L}_{\tiny{weight-CE}} = \sum_{i = 1}^{N} \alpha_{i} \mathrm{CE}(\mathbf{\hat{y}}_{i}, \mathbf{y}_{i}), \label{Loss_Weight-CE}
    \end{split}
\end{align}
where $\alpha$ is the distance-based weighting map. The weight of each airway voxel depends on the Euclidean distance to the centerline, which is defined as:
\begin{align}
\alpha_{i}=\left\{\begin{array}{ll}
- \lambda_{\mathrm{fg}}\mathrm{log}(\frac{\mathrm{dc}_{i}}{\mathrm{dc}_{max}} + \varepsilon ),& \mathbf{y}_{i} = 1,\\
1,& \mathbf{y}_{i} = 0, \label{Loss_Weight-CE_alpha}
\end{array}\right.
\end{align}
where the $\mathrm{dc}_{i}$ is the shortest Euclidean distance from its location to the centerline, $\mathrm{dc}_{max}$ is the maximum of $\mathrm{dc}_{i}$ in one sample. $\lambda_{\mathrm{fg}}$ is the weighting factor and $\varepsilon$ is a small positive number to avoid the numerical error. The topo\_completeness in \fig{\ref{fig::tps}} shows the weight profiles of the main bronchus and peripheral bronchus, which is consistent to \eqn{\eqref{Loss_Weight-CE_alpha}}. Finally, The topology completeness loss $\mathcal{L}_{topo\_com}$ is composed of $\mathcal{L}_{\tiny{tversky}}$ and $\mathcal{L}_{\tiny{weight-CE}}$, i.e., $\mathcal{L}_{topo\_com} = \mathcal{L}_{\tiny{tversky}} + \mathcal{L}_{\tiny{weight-CE}}$, which is designed to adaptively pay more attention to the challenging regions.

\textbf{Topology Correctness}: Previous work \cite{Bruijne_3DUnet-Graph} observed that enhancing the topology completeness of the pulmonary airway may decrease the overlap based segmentation accuracy. The underlying cause of this phenomenon is over-segmenting the airway. Nevertheless, some of these errors can be attributed to the annotation difficulty in keeping consistency and completeness on airways of varying sizes. Although some post-processing techniques \cite{krahenbuhl_Efficient_CRF} could alleviate this problem, they add extra trivial parameters tuning tasks that reduce the efficiency of the models. 

In this work, we propose a topology correctness loss function termed as $\mathcal{L}_{topo\_cor}$, aiming to achieve a satisfactory trade-off performance on both 
topology completeness and correctness. The $\mathcal{L}_{topo\_cor}$ is embedded in the whole end-to-end training procedure with the simple warm up technique. The Label Space $\mathcal{Y}$ contains the positive examples: $\mathcal{Y^{+}}$, and the negative examples: $\mathcal{Y^{-}}$. The results of the model prediction are categorized into four components: true positive (TP), true negative (TN), False Positive (FP), and False Negative (FN). $\mathcal{Y^{+}}$ contains the TP and FN, and $\mathcal{Y^{-}}$ contains 
the TN and FP. The concrete definitions that combine with the model can be seen as follows:
\begin{equation}
\begin{aligned}
    \quad \;  \mathrm{TP} &= \sum_{i \in \mathcal{Y^{+}}} \mathbbm{1} [\mathcal{F}(\Theta; \mathbf{x}_{i}) \geqslant \mathrm{T}]\\
                &= \sum_{i \in \mathcal{Y^{+}}} 1 - \mathcal{L}_{zero\_one}(\mathcal{F}(\Theta; \mathbf{x}_{i}, \mathrm{T}), \mathbf{y}_{i})\\
    \mathrm{FP} &= \sum_{i \in \mathcal{Y^{-}}} \mathbbm{1} [\mathcal{F}(\Theta; \mathbf{x}_{i}) \geqslant \mathrm{T}]\\
                &= \sum_{i \in \mathcal{Y^{-}}} \mathcal{L}_{zero\_one}(\mathcal{F}(\Theta; \mathbf{x}_{i}, \mathrm{T}), \mathbf{y}_{i}),\\
\end{aligned}
\end{equation}
where the $\mathcal{L}_{zero\_one}$ denotes the Zero-One loss function, and the TN and FN can be easily defined, similar to TP and FP. As shown in \fig{\ref{fig::tps}}, the $\mathcal{L}_{topo\_cor}$ aims to maximize the area under the PR curve (AUCPR), which guarantees satisfactory performance on both topology completeness and correctness. The approximation of AUCPR is obtained by summing over a series of precision at fixed recall values, which is equal to the sum over recall at fixed precision. The precision is defined as $\mathrm{P = [TP / (TP + FP)]}$ and the recall is $ \mathrm{R = [TP / (TP + FN)]}$. We first define the preliminary optimization problem: Maximize the precision at the fixed recall value of $\delta$, denoted by $\mathrm{P@R}_{\delta}$.

\begin{align}
    \begin{split}
    \mathrm{P@R}_{\delta} &= \max\limits_{\mathcal{F}_{\Theta}}\mathrm{P}\\
    &s.t.\ \mathrm{R} \geq \delta\\
    \Leftrightarrow &= \max\limits_{\mathcal{F}_{\Theta}}\frac{\mathrm{TP}}{\mathrm{TP + FP}}\\
    &s.t.\ \frac{\mathrm{TP}}{|\mathcal{Y^{+}}|} \geq \delta,\label{eq::PR}
    \end{split}
\end{align}
where $\mathcal{Y^{+}}$ is the total account of the positive examples. Since the $\mathcal{L}_{zero\_one}$ is non-convex and hard to be optimized by SGD algorithms, we choose the hinge loss as a natural replacement, and the $\mathrm{TP}$ and $\mathrm{FP}$ should be bounded as follows:
\begin{equation}
    \begin{aligned}
    \quad \; \mathrm{TP} &= \sum_{i \in \mathcal{Y^{+}}} 1 - \mathcal{L}_{zero\_one}(\mathcal{F}(\Theta; \mathbf{x}_{i}, \mathrm{T}), \mathbf{y}_{i})\\
                &\geq \sum_{i \in \mathcal{Y^{+}}} 1 - \mathcal{L}_{hinge}(\mathcal{F}(\Theta; \mathbf{x}_{i}, \mathrm{T}), \mathbf{y}_{i})\\
                &= \mathrm{TP}^{l},\\
    \mathrm{FP}  &= \sum_{i \in \mathcal{Y^{-}}} \mathcal{L}_{zero\_one}(\mathcal{F}(\Theta; \mathbf{x}_{i}, \mathrm{T}), \mathbf{y}_{i}),\\
                &\leq \sum_{i \in \mathcal{Y^{-}}} \mathcal{L}_{hinge}(\mathcal{F}(\Theta; \mathbf{x}_{i}, \mathrm{T}), \mathbf{y}_{i})\\
                &= \mathrm{FP}^{u},
\end{aligned}
\end{equation}
where the $\mathcal{L}_{hinge}(\mathcal{F}(\Theta; \mathbf{x}_{i}, \mathrm{T}), \mathbf{y}_{i}) = max\{0, 1 - \mathbf{y}_{i}[\mathcal{F}_{\Theta}(\mathbf{x}_{i}) - \mathrm{T}]\}$, $\mathrm{TP}^{l}$ is the lower bound of the $\mathrm{TP}$, and $\mathrm{FP}^{u}$ is the upper bound of the $\mathrm{FP}$. Then the objective of the \eqn{\eqref{eq::PR}} can be transformed as follows:
\begin{align}
    \begin{split}
    \widehat{\mathrm{P@R}}_{\delta} &= \min\limits_{\mathcal{F}_{\Theta}}-\frac{\mathrm{\delta\,|\mathcal{Y^{+}}|}}
    {\delta\,|\mathcal{Y^{+}}| + \mathrm{FP}^{u}}\\
    &s.t.\ \frac{\mathrm{TP}^{l}}{|\mathcal{Y^{+}}|} - \delta \geq 0\label{eq::PR_Transformed}
    \end{split}
\end{align}
We apply the Lagrange multiplier to obtain the equivalent objective function:
\begin{align}
    \begin{split}
    \widehat{\mathrm{P@R}}_{\delta} &= \min\limits_{\mathcal{F}_{\Theta}}\max\limits_{\nu \geq 0}
    -\frac{\mathrm{\delta\,|\mathcal{Y^{+}}|}}{\delta\,|\mathcal{Y^{+}}| + \mathrm{FP}^{u}} + 
    \nu (\frac{\mathrm{TP}^{l}}{|\mathcal{Y^{+}}|} - \delta),\label{eq::PR_Transformed_Lagrang}
    \end{split}
\end{align}
where $\nu$ is a Lagrangian multiplier and \eqn{\eqref{eq::PR_Transformed_Lagrang}} can be solved by SGD algorithm, iteratively updating $\Theta$ and $\nu$. Since we have 
derived the objective function of $\widehat{\mathrm{P@R}}_{\delta}$, the $\mathcal{L}_{topo\_cor}$ aims to enhance topology correctness while maintaining the topology completeness. Hence, the final $\mathcal{L}_{topo\_cor}$ is defined by:
\begin{align}
    \begin{split}
    \mathcal{L}_{topo\_cor} = \min\limits_{\mathcal{F}_{\Theta}}-\sum_{k = 1}^{m}
    (\delta_k - \delta_{k-1})\widehat{\mathrm{P@R}}_{\delta_k},\\
    \end{split}
\end{align}
where $\delta_{k} = \delta_{0} + \frac{(1 - \delta_{0})k}{m}$, $\delta_{0}$ is the positive class prior. The $\mathcal{L}_{topo\_cor}$ cooperating with the $\mathcal{L}_{topo\_com}$ aids in preserving the topology structure of the pulmonary airway, the TPS loss function is finally defined as:
\begin{align}
    \begin{split}
    \mathcal{L}_{TPS} = \lambda_{1}\mathcal{L}_{topo\_com} + \lambda_{2}\mathcal{L}_{topo\_cor}, 
    \end{split}
\end{align}
where $\lambda_{1}$ and $\lambda_{2}$ are hyperparameters to balance the loss terms. We empirically set $\lambda_{1} = \lambda_{2} = 1$ in this work.

\subsection{Convolutional Distance Transform}

Recent studies have demonstrated that introducing distance transform maps into CNNs could boost the performance of medical image segmentation~\cite{kervadec2019boundary,xue2020shape,wang2020deep,ma2020learning}. Compared with the binary segmentation masks, the distance transform maps can highlight the local structure features by measuring the distance between voxels to the nearest boundary. As shown in \fig{\ref{fig::dist_trans}}, the local breakage can be easily observed by this map even if it has slight impacts on overlap based metrics. 
However, these methods mainly suffer two drawbacks: 1) It is intractable to directly regress the SDM, especially for the complicated airway tree structure. 2) The commonly used distance transform is non-differentiable, therefore it is difficult to unify the distance transform map with the network in an end-to-end fashion.
In this paper, we introduce the Convolutional Distance Transform (CDT) module that functions on the probability output of the backbone. The main idea of the CDT is to build a differentiable distance transform as a loss function, collaborating with the TPS to preserve the topology structure of the pulmonary airway. 
The core challenge of CDT is to determine the foreground voxels in the probability map and then calculate the shortest distance from all inside foreground voxels to their boundary. To tackle such a problem, we design a sequential procedure to construct CDT, consisting of the categorical reparameterization, soft boundary extraction, and kernel-based distance transform. Given that each component of CDT is differentiable, therefore the overall gradient flow is guaranteed. 

Firstly, the gumbel-softmax~\cite{jang2016categorical} is used as an alternative to discarding the $\mathrm{argmax}$ gate. The logits of probability map $\mathbf{\hat{y}}$ can be converted into the binary result and the gradient map through gumbel-softmax is computed identically to the ones through softmax. Specifically, $\mathbf{\hat{y}} = [\mathbf{\hat{y}^{1}};\mathbf{\hat{y}^{2}}]$ contains two maps, where $\mathbf{\hat{y}^{1}}$ and $\mathbf{\hat{y}^{2}}$ denote the probability map of background and foreground respectively. The gumbel-softmax distribution adopts softmax as a continuous relaxation to $\mathrm{argmax}$, hence, we can acquire the binary result $\mathbf{\hat{z}}$ via the differentiable equation: 
\begin{align}
    \begin{split}
    \mathbf{\hat{z}^{i}} = \frac{\mathrm{exp}(\mathrm{log}((\mathbf{\hat{y}^{i}}) + \mathbf{{g}^{i}})/\tau)}
    {\sum_{j=1}^{2}\mathrm{exp}(\mathrm{log}((\mathbf{\hat{y}^{j}}) + \mathbf{{g}^{j}})/\tau)},\;\mathrm{for\;} i = 1, 2.\label{eq::Gumbel-Softmax}
    \end{split}
\end{align}
where $\mathbf{\hat{g}^{i}}$ represents the standard Gumbel distribution: $\mathbf{{g}} = -\mathrm{log}(-\mathrm{log}(\mathbf{{u}}))$ with $\mathbf{{u}}$ sampled from a uniform distribution, i.e., $\mathbf{{u}} \sim Unif[0,1]$. $\tau$ is the temperature parameter that controls the discreteness of the binary result $\mathbf{\hat{z}}$. When $\tau$ becomes closer to 0, the samples from the Gumbel Softmax distribution become almost the same as the one-hot argmax output. We set $\tau=0.1$ in practical use, which is effective enough for the accuracy of experiments.

Secondly, the binary result $\mathbf{\hat{z}}$ is composed of the foreground map $\mathbf{\hat{z}}^{\mathrm{fg}} = \mathbf{\hat{z}}^{2} $ and the background map $\mathbf{\hat{z}}^{\mathrm{bg}} =  \mathbf{\hat{z}}^{1}$. We simulate the morphological erosion and dilation operations via the $\mathrm{max\_pooling}$ function. These operations are key components to extracting the boundary $\mathbf{\Phi}$ from the foreground map $\mathbf{\hat{z}}^{\mathrm{fg}}$:
\begin{align}
    \begin{split}
    \mathbf{\Phi} = [\mathrm{soft\_dilation}(\mathbf{\hat{z}}^{\mathrm{fg}}) - \mathrm{soft\_erosion}(\mathbf{\hat{z}}^{\mathrm{fg}})] \odot \mathbf{y}, \label{eq::SoftBoundary}
    \end{split}
\end{align}
where  $\mathrm{soft\_erosion}$ and $\mathrm{soft\_dilation}$ follow the same implementation of clDice~\cite{shit2021cldice}.
\begin{figure}[t]
\centering{\includegraphics[width=1.0\linewidth]{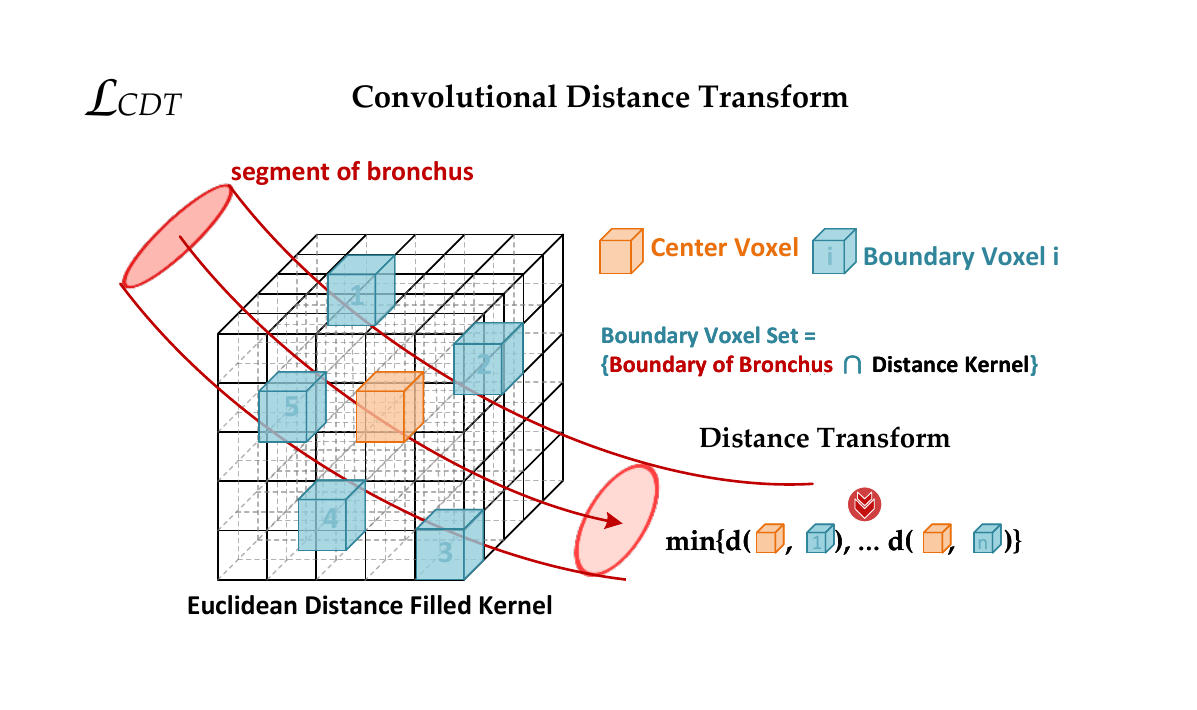}}
\caption{The illustration of CDT module. The shortest distance between the center voxel and the boundary voxel set is calculated based on the Euclidean distance filled kernel.}
\label{fig::cdt}
\end{figure}

Thirdly, we introduce the kernel-based method to solve the distance transform of the $\mathbf{\hat{z}}^{\mathrm{fg}}$. The key step is to solve the shortest distance from each voxel in the $\mathbf{\hat{z}}^{\mathrm{fg}}$ to the boundary $\bm{\Phi}(\varphi)$, which can be defined as:
\begin{align}
    \begin{split}
    \mathbf{Dist}(\mathbf{\hat{z}}^{\mathrm{fg}}) = \min_{\forall \varphi \in \bm{\Phi}} \mathrm{d}(\mathbf{\hat{z}}^{\mathrm{fg}}, \varphi), 
    \end{split}
\end{align}
where $d(\cdot,\cdot)$ denotes the Euclidean distance. Take one voxel $p \in \mathbf{\hat{z}}^{\mathrm{fg}}$ and $\varphi \in \bm{\Phi}$ as example, 
$\mathrm{d}(p,\varphi): \mathbb{R}^{3} \rightarrow \mathbb{R}^{+}_{0}$ can be defined as: 
\begin{align}
    \begin{split}
    \mathrm{d}(p,\varphi) = \mathrm{d}(p - \varphi, 0) = \mathrm{d}(p - \varphi) = {\lVert p - \varphi \rVert}_{2}
    \end{split}
\end{align}
The question now is transformed to distinguish the shortest distance from the distance set via differentiable computation. 
The distance set for a singe voxel $p$ contains:$\{\mathrm{d}_{1}, \mathrm{d}_{2}, ... \mathrm{d}_{n}\}$ = 
$\{\mathrm{d}(p,\varphi_{1}), \mathrm{d}(p,\varphi_{2}), ...\mathrm{d}(p,\varphi_{n})\}$. 
A convolutional distance-kernel is initialized with the Euclidean distance value to the centroid, vividly seen in \fig{\ref{fig::cdt}}, the size of distance-kernel is set to $31 \times 31 \times 31$ guaranteeing the airway voxels could touch the boundary. Next, we use the property of the log-sum-exponential~\cite{cook2011basic} to reformulate the problem:
\begin{align}
    \begin{split}
    \mathit{LSE}_{\beta}(\mathrm{d}_{1},...,\mathrm{d}_{n}) = \frac{1}{\beta}\mathrm{log}(\mathrm{exp}(\beta\mathrm{d}_{1}) + ... + \mathrm{exp}(\beta\mathrm{d}_{n}))
    \end{split}
\end{align}
Without loss of generality, we can assume that \{$\mathrm{d}_{1} = \mathrm{d}_{2} = ... = \mathrm{d}_{k} < \mathrm{d}_{k+1} \leqslant ... \,\mathrm{d}_{n}$\}, and 
\textit{Lemma 1} derives the minimum distance via log-sum-exponential.
\begin{lemma}
    $\min\{\mathrm{d}_{1}, \mathrm{d}_{2}, ... \mathrm{d}_{n}\} = \lim\limits_{\beta \to -\infty} \mathit{LSE}_{\beta}(\mathrm{d}_{1},...,\mathrm{d}_{n})$ \label{eq::LSELemma}
\end{lemma}
Detailed proof can be seen in Appendix. Finally, we conclude the proposed CDT as follows: 
\begin{align}
    \begin{split}
        \mathbf{Dist}(\mathbf{\hat{z}}^{\mathrm{fg}}) &= 
        \lim\limits_{\beta \to -\infty} \frac{1}{\beta}\mathrm{log}(\bm{\Phi}(\varphi) \cdot \mathrm{exp}(\beta \mathrm{d}(\mathbf{\hat{z}}^{\mathrm{fg}} - \varphi))),\\
        &= \lim\limits_{\beta \to -\infty} \frac{1}{\beta}\mathrm{log}(\bm{\Phi}(\varphi) \ast \mathrm{exp}(\beta \mathrm{d}(\mathbf{\hat{z}}^{\mathrm{fg}}))), \\
        &= \lim\limits_{\gamma \to -0} \gamma \mathrm{log}(\bm{\Phi}(\varphi) \ast \mathrm{exp}(\frac{1}{\gamma}\mathrm{d}(\mathbf{\hat{z}}^{\mathrm{fg}}))),\label{eq::CDT_Final_Formula} 
    \end{split}
\end{align}
where the $\ast$ is the convolutional operator, and $\gamma = \frac{1}{\beta}$. We set $\gamma = -0.3$ in experiments that enable the smooth approximation 
of the CDT. Since all components are differentiable, the proposed CDT could be integrated as a differentiable distance transform layer into the end-to-end training framework. Given that the distance transform map of the probability map is obtained, we further design a distance map loss that penalizes the false predictions from the global perspective:  
\begin{align}
    \begin{split}
    \mathcal{L}_{CDT} = \omega \cdot {\left\lVert \mathbf{Dist}(\mathbf{y}) - \mathbf{Dist}(\mathbf{\hat{z}}^{\mathrm{fg}})\right\rVert }_{2}, \label{eq::Loss_CDT}
    \end{split}
\end{align}
where the $\mathbf{\omega}$ is a weighting balance to handle the imbalance between the foreground and the background.
\begin{align}
\mathbf{\omega}=\left\{\begin{array}{ll}
\frac{N_{\mathrm{bg}}}{N_{\mathrm{fg}}},&\mathrm{if\;foreground},\\
1,& \mathrm{otherwise}.
\end{array}\right.
\end{align}
where the $N_{\mathrm{fg}}$ and $N_{\mathrm{bg}}$ denote the number of voxels in the foreground and background, respectively. In conclusion, the proposed DTPDT network owns two critical modules, the TPS and CDT. 
These two modules cooperate with each other to enhance the segmentation performance on pulmonary airway.
\section{Experiments}
\subsection{Datasets}
Two public datasets are used for evaluation in our work. 1) \textbf{The EXACT'09 Challenge} \cite{EXACT09}. It provides 20 CT scans for training and 20 CT scans for testing, however no airway annotation is publicly available. All CT scans share an axial size of 512$\times$512, with a spatial resolution ranging from 0.5mm to 0.78mm. The quantity of the axial slices varies from 157 to 764 and their slice thickness ranges between 0.45mm to 1.0mm.2) \textbf{The Binary Airway Segmentation Dataset (BAS)} \cite{qin2020learning}. The BAS contains 90 CT scans (70 from LIDC\cite{armato2011lung} and 20 from EXACT'09) with airway annotation. The spatial resolution ranges from 0.5mm to 0.82 mm and the slice thickness ranges from 0.5mm to 1.0 mm. For the BAS dataset, we randomly split the 90 CT scans into the training set (50 scans), validation set (20 scans), and test set (20 scans). The model trained on the BAS dataset was used to evaluate on the test set of the EXACT'09 challenge. For a fair comparison, the resulting binary segmentations were submitted to the organizers who sent the quantitative evaluations back.

\subsection{Implementation Details}
\noindent \textbf{Network Configuration and Data Preprocessing}: As shown in the \fig{\ref{MainFramework}}, 3D UNet was chosen as the backbone with a slight modification, 
each block in the encoder or decoder contains two convolutional layers followed by Instance Normalization\cite{ulyanov2016instance} and pReLU\cite{he2015delving}. Initial channel number was set to 32. During the preprocessing procedure, we clamped the voxel values to [-1000,600] Hounsfiled Unit, normalized them into [0, 255], 
and cropped the lung field to remove unrelated background regions. 

\noindent \textbf{Optimization Procedure}: 
We adopted a large input size of 128 $\times$ 224 $\times$ 304 CT cubes densely cropped near airways and chose a batch size of 1 in the training phase. On-the-fly data augmentation included the random horizontal flipping and random rotation between [-$10^{\degree}$, $10^{\degree}$].  Adam optimizer was used with the initial learning rate of 0.002. The total epoch was set to 60, the first ten epochs were used as warm up solution. The learning rate was divided by 10 in the 50$^{th}$ epoch. 
We chose $\lambda_{\mathrm{fg}} = 10$, $\delta = 0.8$, and $m = 10$ in the TPS module via experimental results. The detailed training optimization of the proposed DTPDT framework can refer to the \textcolor{Algorithm}{Algorithm. \ref{training procedure}}. During the testing phase, we performed the sliding window prediction with stride 48. The results were averaged on the overlapping regions and a threshold ($th = 0.5$) function was used to obtain the final binary segmentation results. The proposed model took around 12 seconds to predict a sub-volume CT cube with the size of 128 $\times$ 224 $\times$ 304. We adopted the PyTorch framework to implement all experiments, which were executed on a linux workstation with Intel Xeon Gold 5218 CPU @ 2.30 HZ, 128 GB RAM, and 2 NVIDIA Geforce RTX 3090 GPUs.

\noindent \textbf{Evaluation Metrics}: For EXACT'09 Challenge, the full test results evaluated by the organizers were preserved, as seen in \tab{\ref{Table_EXACT_Result}}. It can be noticed that the metrics used by EXACT'09 can be categorized into the topology and overlap based accuracy measurements. Therefore, we adopted the tree length detected rate (TD, \%) and branch detected rate (BD, \%) to evaluate the topological completeness and connectedness. In addition, we used the DSC to assess the overlap based accuracy.

\begin{algorithm}[!t]
\caption{Main optimization procedure of the DTPDT.}
\SetAlgoLined
\KwIn{input $\mathbf{x}$, label $\mathbf{y}$, current epoch $T_{c}$, warm up epoch $T_{w}$. Initialize the backbone $\mathcal{F}_{\Theta}$.} 
    \While{\textit{not converged}}{
    Acquire likelihood map $\mathbf{\hat{y}}$ from $\mathcal{F}_{\Theta}$, $\mathbf{\hat{y}} = \mathcal{F}_{\Theta}(\mathbf{x})$. \par
    if $T_{c}$ $<$ $T_{w}$: \par
    \hskip\algorithmicindent Compute topology completeness loss, $\mathcal{L}_{topo\_com}(\mathbf{\hat{y}}, \mathbf{y})$. \par 
    \hskip\algorithmicindent Update $\Theta$ with $\mathcal{L}_{topo\_com}$.\par
    else: \par
    \hskip\algorithmicindent Compute  $\mathcal{L}_{topo\_com}(\mathbf{\hat{y}}, \mathbf{y})$. \par 
    \hskip\algorithmicindent Compute $\mathcal{L}_{topo\_cor}(\mathbf{\hat{y}}, \mathbf{y})$. \par
    \hskip\algorithmicindent Use Gumbel-Softmax to obtain $\mathbf{\hat{z}}^{\mathrm{fg}}$ from $\mathbf{\hat{y}}$. \par  
    \hskip\algorithmicindent Use soft morphological operations to get $\Phi$ from $\mathbf{\hat{z}}^{\mathrm{fg}}$. \par  
    \hskip\algorithmicindent Calculate the distance transform of $\mathbf{y}$ and $\mathbf{\hat{z}}^{\mathrm{fg}}$ via the \par 
    \hskip\algorithmicindent CDT module, $\mathbf{Dist(\mathbf{y})}$ and $\mathbf{Dist(\mathbf{\hat{y}})}$. Then compute \par 
    \hskip\algorithmicindent distance map loss, $\mathcal{L}_{CDT}(\mathbf{Dist(\mathbf{\hat{y}})}, \mathbf{Dist(\mathbf{y})})$. \par 
    \hskip\algorithmicindent Update $\Theta$ with $\mathcal{L}_{topo\_com}$ + $\mathcal{L}_{topo\_cor}$ + $\mathcal{L}_{CDT}$.\par
}
\label{training procedure}
\end{algorithm}

\subsection{Quantitative Results Analysis}
\makeatletter
\def\hlinew#1{%
\noalign{\ifnum0=`}\fi\hrule \@height #1 \futurelet
\reserved@a\@xhline}
\makeatother
\begin{table}[!t]
\renewcommand\arraystretch{1.4}
\centering
\caption{Comparison in the EXACT-09 Dataset. The results are reported in the format of mean $\pm$ standard deviation. All results are acquired from the official organization. For simplicity, `BD' represents Branch Detected, `TD' represents Tree Length Detected, and `FPR' represents False Positive Rate.}\label{Table_EXACT_Result}
\scalebox{0.95}{
\begin{threeparttable}
\begin{tabular}{lccccccc}
\hlinew{1pt}
Method &  BD (\%) $\uparrow$  & TD (\%) $\uparrow$ & FPR(\%) $\downarrow$ \\ \hline
Neko\tnote{$\dag$}  & 35.5 $\pm$ 8.2  & 30.4 $\pm$ 7.4 & \underline{0.89 $\pm$ 1.78} \\
Murphy \etal\cite{nardelli2015optimizing} & 41.6 $\pm$ 9.0 &  36.5 $\pm$ 7.6 & \textbf{0.71 $\pm$ 1.67} \\
Inoue \etal\cite{inoue2013robust}  & 79.6 $\pm$ 13.5  & \textbf{79.9 $\pm$ 12.1} &11.92 $\pm$ 13.16 \\
Xu \etal\cite{xu2015hybrid} & 51.1 $\pm$ 10.9 & 43.9 $\pm$ 9.6 & 6.78 $\pm$ 26.60 \\
MISLAB\tnote{$\dag$}  & 42.9 $\pm$ 9.6  & 37.5 $\pm$ 7.1 & \underline{0.89 $\pm$ 1.64} \\
Smistad\etal\cite{smistad2014gpu} & 31.3 $\pm$ 10.4 & 27.4 $\pm$ 9.6 &3.60 $\pm$ 3.37 \\ \hline
Qin \etal (\textit{th} = 0.8)\cite{qin2020learning} & 68.8 $\pm$ 13.4 & 62.6 $\pm$ 12.7 & 1.28 $\pm$ 1.29 \\
Qin \etal (\textit{th} = 0.1)\cite{qin2020learning} & \underline{82.0 $\pm$ 9.9} & 79.4 $\pm$ 10.0 & 9.71 $\pm$ 5.59 \\
Qin \etal (\textit{th} = 0.5)\cite{qin2020learning} & 76.7 $\pm$ 11.5 & 72.7 $\pm$ 11.6 &3.65 $\pm$ 2.86 \\
Zheng \etal\cite{zheng2021alleviating} & 80.5 $\pm$ 12.5 & 79.0 $\pm$ 11.1 & 5.79 $\pm$ 4.25 \\
Yu \etal\cite{yu2022break} & 78.3 $\pm$ 14.9 & 77.1 $\pm$ 13.6 & 4.27 $\pm$ 2.79 \\
\hline
DTPDT & \textbf{82.1 $\pm$ 10.6} &  \underline{79.6 $\pm$ 9.5} & 6.32 $\pm$ 4.55 \\ 
\hlinew{1pt}
\end{tabular}
\begin{tablenotes}
\scriptsize
\item[$\dag$] Directly use the name of the participant team.
\end{tablenotes}
\end{threeparttable}}
\end{table}
The results of the EXACT'09 Challenge test data are reported in the \tab{\ref{Table_EXACT_Result}}. It is observed that the proposed DTPDT network achieved the highest performance on the metric of BD (82.1\%) and the second-highest performance on TD (79.6\%). Compared with other methods~\cite{qin2020learning,inoue2013robust}, our approach improved the topology completeness without sacrificing too much of the topology correctness. Specifically, Inoue \etal \cite{isensee2021nnu} designed the multi-stages framework for airway extraction. Although they achieved the highest TD (79.9\%), the side-effect was also conspicuous, as the FPR reaches 11.92\%. Our approach could obtain similar high TD (79.6\%) while maintaining the modest FPR (6.32\%). Further, on the metric of branch detected rate, we achieved 82.1\% BD, which had 2.5 percentage points higher than the method of Inoue. This observation reveals that the proposed DTPDT network had truly detected more different bronchi rather than extending the length of partial bronchi. Qin \etal\cite{qin2020learning} performed another important experiment under different thresholds to generate the final segmentation mask. It is observed that the BD and TD were increased while the FPR suffered a noticeable degeneration along with the relaxation of the threshold. This indicates that it is challenging to  balance the recall and the precision via tunning the hyper-parameters. The proposed DTPDT network outputs the binary airway segmentation with the threshold set to 0.5 (commonly used in binary classification tasks). Compared with the 82.0\% BD and 79.4\% TD achieved by Qin under the threshold of 0.1, we obtained a higher performance 
of 82.1\% BD and 79.6\% TD. In addition, our FPR was lower than Qin (6.32\% \textit{v.s.} 9.71\%). This phenomenon collaborates that our method effectively achieves the trade-offs between topology completeness and topology correctness. Zheng \etal\cite{zheng2021alleviating} and Yu \etal\cite{yu2022break} adopted the WingsNet\cite{zheng2021alleviating} as the backbone, designing General Union loss and breakage-sensitive loss to resolve the airway segmentation, respectively. Compared to these methods, our approach revealed better capability in persevering topology structure of the airways (+ 1.6\% BD, + 0.6\% TD compared to Zheng \etal\cite{zheng2021alleviating}, 
+ 3.8\% BD, + 2.5\% TD compared to Yu \etal\cite{yu2022break}) with only a slight increase of the FPR (approximately + 0.5\%). 

The experimental results in BAS dataset, reported in \tab{\ref{Table_LIDC_Result}}, were broadly similar to those in EXACT'09 Challenge test data. 
First, some state-of-the-art airway segmentation algorithms\cite{Bruijne_Juarez_Automatic,Bruijne_3DUnet-Graph,qin2019airwaynet,qin2020learning} 
were re-implemented for comparison. Juarez \etal \cite{Bruijne_Juarez_Automatic} trained a 3D UNet with a compound loss function to automatically segment airways. 
Further, they replaced the bottleneck layer with a Graph Neural Network (GNN) module. The results showed they achieved high DSC performance but 
failed to detect the small bronchi. Qin \etal \;designed the AirwayNet\cite{qin2019airwaynet} to predict the connectivity of airways. With the integration of feature recalibration and attention distillation, the TD and BD could increase from 84.16\% and 78.45\% to 90.89\% and 87.51\%, respectively. We also implemented the nnUNet\cite{isensee2021nnu} since it has established very strong baselines for medical image segmentation tasks. It could achieve satisfactory DSC (93.12\%) while the TD (84.32\%) and BD (85.87\%) are relatively low, which demonstrates the necessity to develop new components for preserving the topological structures.
The representative topology information embedded methods were also taken into consideration in our experiments. LTSP\cite{wu2022ltsp} introduced a long-term slice propagation to capture long-term continuity information. However, the enhancement is limited since the propagation directions had not been exploited to the full. DDT\cite{wang2020deep} combined the CNN with the level-set functions, rephrasing the distance prediction problem as a classification problem based on quantization. Unfortunately, 
the Geometry-aware Refinement (GAR) in DDT is non-differentiable and adds an extra burden to the optimization procedure.

\makeatletter
\def\hlinew#1{%
\noalign{\ifnum0=`}\fi\hrule \@height #1 \futurelet
\reserved@a\@xhline}
\makeatother
\begin{table}[!t]
\renewcommand\arraystretch{1.4}
\centering
\caption{Comparison in the Binary Airway Segmentation dataset. The results are reported in the format of mean $\pm$ standard deviation.}\label{Table_LIDC_Result}
\scalebox{0.8}{
\begin{threeparttable}
\begin{tabular}{@{}lcccc@{}}
\hlinew{1pt}
Method &  TD (\%) $\uparrow$  & BD (\%) $\uparrow$ & DSC (\%) $\uparrow$ & FPR(\%) $\downarrow$ \\ \hline
Juarez \emph{et al.} 2018 \cite{Bruijne_Juarez_Automatic} & 84.12 $\pm$ 9.15 & 74.15 $\pm$ 13.5 & 92.75 $\pm$ 1.97 & 0.017 $\pm$ 0.009 \\
Juarez \emph{et al.} 2019 \cite{Bruijne_3DUnet-Graph} & 84.85 $\pm$ 8.67 & 75.33  $\pm$ 12.4  & \textbf{93.26 $\pm$ 2.25} & \underline{0.014 $\pm$ 0.010} \\
Qin \emph{et al.} 2019 \cite{qin2019airwaynet} & 84.16 $\pm$ 10.4 & 78.45 $\pm$ 9.51 & \underline{93.15 $\pm$ 2.74} & \textbf{0.014 $\pm$ 0.009} \\
Qin \emph{et al.} 2020 \cite{qin2020learning} & 90.89 $\pm$ 5.42 & 87.51 $\pm$ 9.94 & 92.45 $\pm$ 3.05 & 0.035 $\pm$ 0.014 \\
nnUNet~\cite{isensee2021nnu} & 84.34 $\pm$ 7.59 & 85.87 $\pm$ 5.83 & 93.12 $\pm$ 1.16 & 0.018 $\pm$ 0.011 \\ \hline
LTSP~\cite{wu2022ltsp}  & 87.59 $\pm$ 8.71 & 79.83 $\pm$ 11.4 & 92.95 $\pm$ 1.61 & 0.030 $\pm$ 0.017 \\
DDT~\cite{wang2020deep} & 89.27 $\pm$ 7.73 & 88.10 $\pm$ 6.23 & 91.98 $\pm$ 1.42 & 0.042 $\pm$ 0.018 \\
clDice~\cite{shit2021cldice} & 88.45 $\pm$ 7.59 & 86.67 $\pm$ 5.83 & 92.34 $\pm$ 1.04 & 0.021 $\pm$ 0.009 \\
GUL~\cite{zheng2021alleviating} & 89.51 $\pm$ 6.91 & 84.53 $\pm$ 5.18 & 92.78 $\pm$ 1.05 & 0.024 $\pm$ 0.010 \\
Boundary Loss~\cite{kervadec2019boundary}  & 90.45 $\pm$ 8.41 & 88.98 $\pm$ 6.07 & 92.74 $\pm$ 1.41 & 0.047 $\pm$ 0.017 \\
SDM Learning~\cite{xue2020shape}  & 91.43 $\pm$ 11.9 & 89.12 $\pm$ 11.4 & 91.45 $\pm$ 1.25 & 0.043 $\pm$ 0.017 \\ \hline
+ TPS     & \underline{95.44 $\pm$ 3.92} & \underline{90.50 $\pm$ 3.01} & 91.88 $\pm$ 2.09 & 0.062 $\pm$ 0.021 \\
+ CDT & 92.60 $\pm$ 7.60 & 89.38 $\pm$ 6.26 & 92.06 $\pm$ 1.98 & 0.044 $\pm$ 0.020 \\
DTPDT (TPS and CDT) & \textbf{96.52 $\pm$ 3.95} & \textbf{91.50 $\pm$ 2.99} & 91.24 $\pm$ 2.78 & 0.049 $\pm$ 0.018 \\
\hlinew{1pt}
\end{tabular}
\end{threeparttable}}
\end{table}

The clDice loss function\cite{shit2021cldice} was also compared in our experiments, however, the iterative 
pooling operations cannot guarantee the accurate soft skeleton for loss computation, which limited its performance. 
The distance transform methods showed potential in preserving the topology completeness, as the TD and BD reached up to 90.45\% and 
88.98\% with the introduction of boundary loss. The SDM learning\cite{xue2020shape} converted the 
segmentation task into predicting the SDM, which was demonstrated to retain better continuity of shape. This learning strategy 
could boost the performance to 91.43\% TD, 89.12\% BD, and 91.45\% DSC. However, these methods can not explicitly resolve the 
breakage problem and the direct regression of SDM is intractable. Hence, we designed the Convolutional Distance Transform (CDT) to be 
sensitive to the breakage phenomenon and constructed the corresponding loss function to preserve the topological structure correctly. 
The CDT improved the airway segmentation to 92.60\% TD, 89.38\% BD and 92.06\% DSC. 
\subsection{Qualitative Results Analysis}
\begin{figure}[t]
\centering{\includegraphics[width=0.9\linewidth]{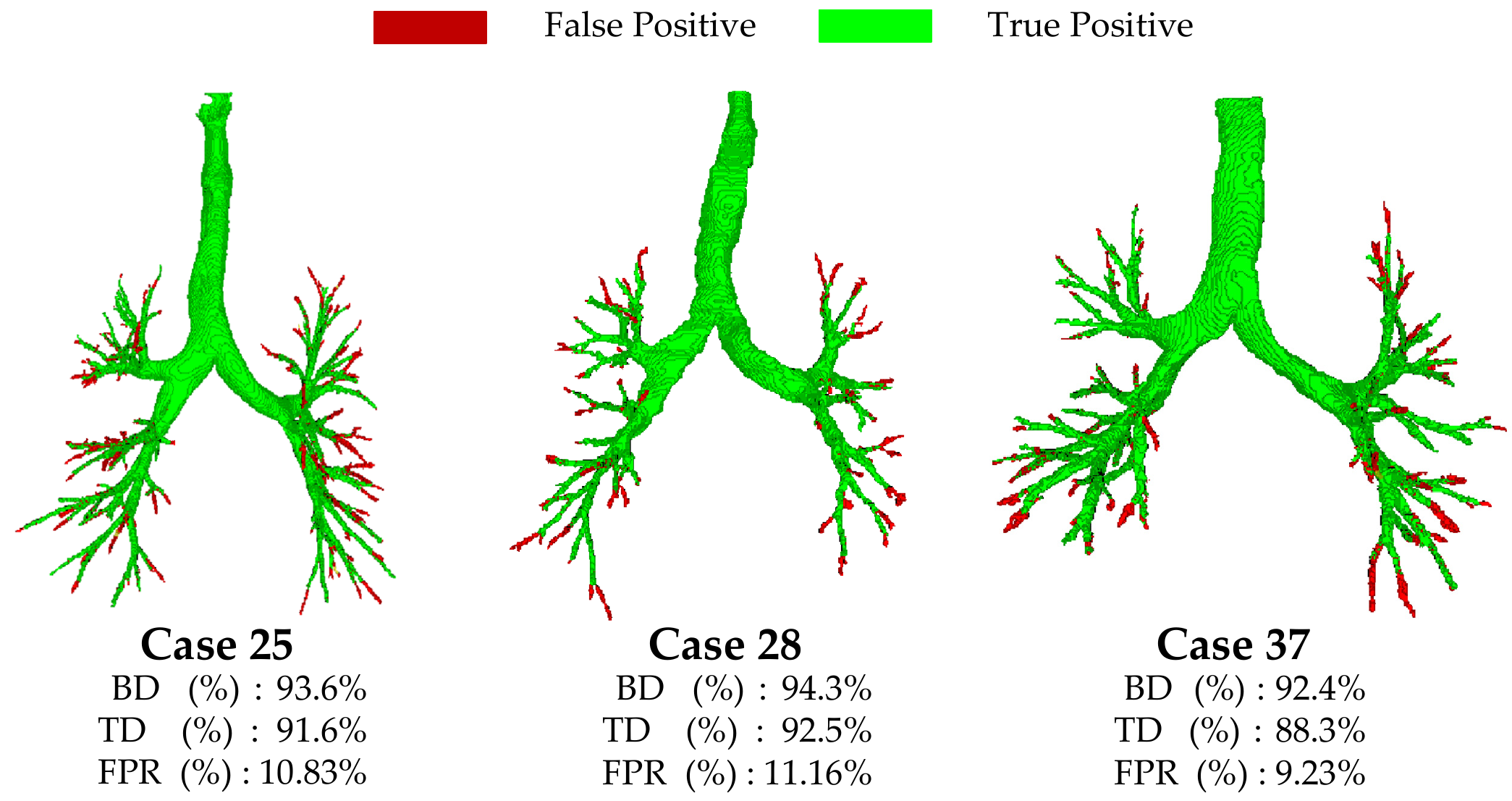}}
\caption{Three cases with the lowest FPR in the EXACT'09 test dataset are presented. The green represents the true positive and the 
red denotes the false positive (best viewed in color). The false positives are almost the distal bronchi. No significant mass of leakage 
is observed.}
\label{fig::EXACT_09_Visualization}
\end{figure}
For the EXACT'09 Challenge test data, we presented three cases with inferior FPR in \fig{\ref{fig::EXACT_09_Visualization}}. It 
can be seen that almost all false positives belong to the distal airways, which may be not annotated by the experts. 
No severe clumps of leakage of the airway were observed in the EXACT'09 results. 

\begin{figure*}[t]
\centering
\centering{\includegraphics[width=0.85\linewidth]{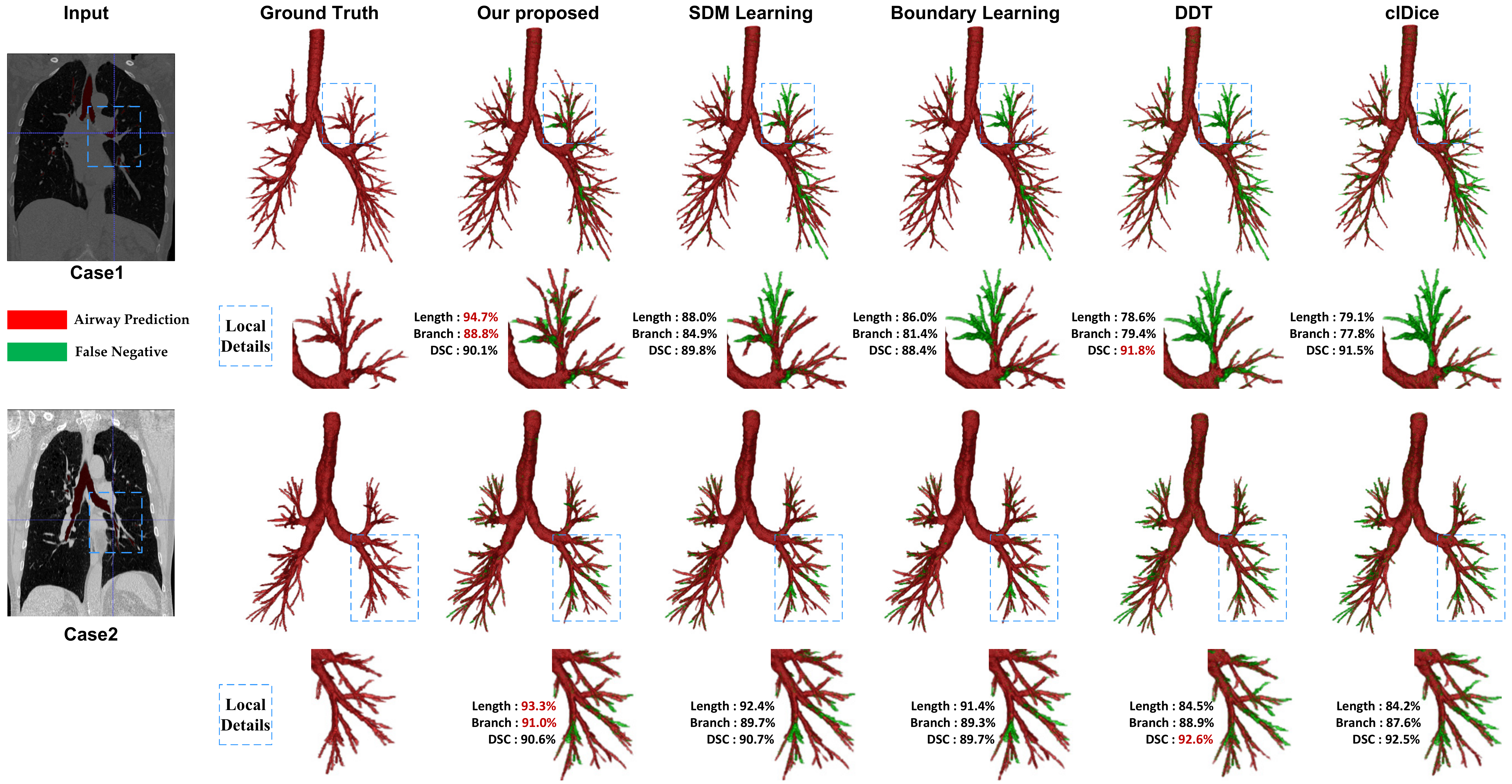}}
\caption{Visualization of the segmentation results of the proposed approach and compared methods. 
Only the largest component of the predictions are rendered in red. The green part reveals the false positives compared to the ground-truth. 
Blue dotted boxes represent local details of airway results. Best viewed in color and magnified. } \label{LIDC_Visualization}
\end{figure*}
For the BAS dataset, compared with other topology information embedded methods, \fig{\ref{LIDC_Visualization}} demonstrated the effectiveness and robustness of our proposed methods. All the airway prediction results were post-processed by the largest component extraction. 
In line with the \tab{\ref{Table_LIDC_Result}}, all methods segmented main bronchi well, however, 
our method preserved the peripheral airway structures more precisely than others, as seen in the local details of Case 1 in the \fig{\ref{fig::EXACT_09_Visualization}}. 
In addition, local details of Case 2 showed that our method can detect as many as possible of distal airways. These observations 
substantiate that DTPDT network could identify the breakage phenomenon with superior sensitivity and preserve the topological structure 
with competitive segmentation accuracy.

\makeatletter
\def\hlinew#1{%
\noalign{\ifnum0=`}\fi\hrule \@height #1 \futurelet
\reserved@a\@xhline}
\makeatother
\begin{table}[!t]
\renewcommand\arraystretch{1.4}
\centering
\caption{Ablation Study of the TPS module on the BAS dataset.}\label{Table_Ablation_in_the_TPS_Module}
\scalebox{0.85}{
\begin{threeparttable}
\begin{tabular}{@{}lcccc@{}}
\hlinew{1pt}
{Method} & \multicolumn{1}{p{1.5cm}<{\centering}}{TD(\%) $\uparrow$} & \multicolumn{1}{p{1.5cm}<{\centering}}{BD(\%) $\uparrow$}  & \multicolumn{1}{p{1.3cm}<{\centering}}{DSC(\%) $\uparrow$} &\multicolumn{1}{p{1.2cm}<{\centering}}{FPR(\%) $\downarrow$}\\ \hline
Topo complete only     & 96.55 $\pm$ 4.53 & 91.25 $\pm$ 4.21 & 86.43 $\pm$ 4.52 & 0.081 $\pm$ 0.028\\
Topo correct only & 91.23 $\pm$ 6.27 & 89.07 $\pm$ 4.91 & 92.82 $\pm$ 1.29 & 0.034 $\pm$ 0.013 \\
TPS (both)    & 95.44 $\pm$ 3.92 & 90.50 $\pm$ 3.01 & 91.88 $\pm$ 2.09 & 0.062 $\pm$ 0.021 \\
\hlinew{1pt}
\end{tabular}
\end{threeparttable}}
\end{table}

\begin{figure}[!t]
\centering{\includegraphics[width=0.8\linewidth]{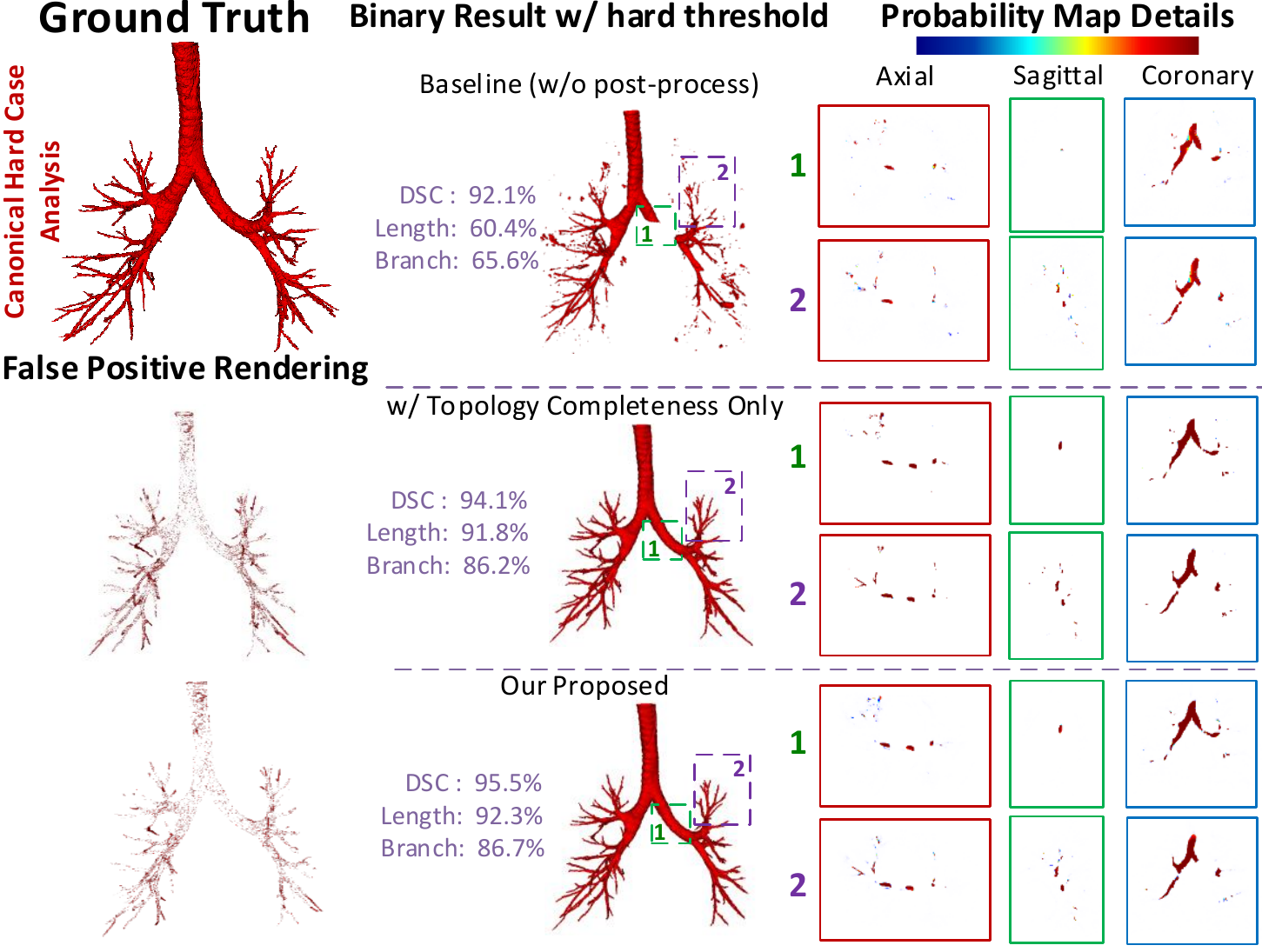}}
\caption{Illustration of binary airway prediction and probability maps obtained by different methods. The baseline method 
generates the skew likelihood map that peripheral airways share lower confidence probability than the principal bronchi, as seen in the 
probability map details of three planes. This problem can be alleviated by the topology information embedded methods. Further, 
false positives are reduced by our proposed method compared to the methods with only topology completeness. }
\label{Triplet_control_experiment}
\end{figure}

\section{Discussions}
\subsection{Impacts of Individual Modules}
We first conduct the ablation study on the proposed method. \tab{\ref{Table_Ablation_in_the_TPS_Module}} reports the impacts of individual modules in TPS. Despite the TD and BD could be improved to 96.55\% and 91.25\%, the DSC and FPR dropped to 86.43\% and 0.081\% if only used the topology completeness loss function. Under the common Dice loss setting, the introduction of the topology correctness objective function revealed its capability to improve the segmentation accuracy. The TPS module combined them to acquire a good trade-off between topology completeness and correctness, which can achieve the 95.44\% TD, 90.50\% BD, 91.88\% DSC, and 0.062\% FPR. The Convolutional Distance Transform (CDT) is designed to be sensitive to the breakage phenomenon and constructed the corresponding loss function to preserve the topological structure correctly. \tab{\ref{Table_LIDC_Result}} also indicates that CDT improved the airway segmentation to 92.60\% TD, 89.38\% BD, 92.06\% DSC, and 0.044\% FPR. The probability map of three profiles in \fig{\ref{Triplet_control_experiment}} vividly revealed that the proposed method can assign proper likelihoods for within-class imbalanced airways to alleviate the breakage problems. Further, compared with the methods with only topology completeness objective function, the false positives are reduced by DTPDT network, as shown in the false positive rendering in \fig{\ref{Triplet_control_experiment}}. 

\subsection{Rethinking the Topology Priors in Deep Learning Era}
This work designed differentiable modules, TPS and CDT, to improve the topology correctness and completeness of airway segmentation. The topology/geometry priors, including the local smoothness, connectivity, and other issues, have been widely investigated in medical imaging computing since segmenting objects while preserving their global structures are important for clinical diagnosis. Previous works have formulated these issues as implicit or explicit constraints during the model optimization. Thanks to the development of deep learning and the increasing size of clinical data, we can easily train a neural network via minimizing the differences between predictions and the ground-truth. However, the loss term should be differentiable since the stochastic optimization and back-propagation are deployed to update the parameters of models. DSC metric (and its variants) is differentiable and plays an important role in segmentation tasks since it can be used for both model training and evaluation. 
Despite the high voxel-wise accuracy achieved by DSC objective function in airway segmentation, it is still prone to structural errors, such as missing small bronchi and breaking thin connections. The non-differentiable metrics including BD and TD are not friendly to deep learning models. Both CDT and TPS are implicit loss terms to improve the topology performance since the best supervision signals to characterize the topology priors are still far away from solved. More specifically, the breakage problem is handled by CDT in this work. However, a generalized representation of segmentation mask is favorable to cover multiple shapes rather than the tubular structures only. 

\subsection{More Challenges and Future of Airway Segmentation}
Both pulmonary disease assessment and endobronchial intervention require accurate airway segmentation for quantitative measurements of bronchial features. Compared with previous works, the proposed method improves the topology accuracy via balancing the recall/precision metrics and reducing the breakages of small bronchi. This can benefit the accurate counting of airway branches and the navigation to reach the peripheral targets. However, the false positives, as shown in \fig{\ref{Triplet_control_experiment}}, are still not avoidable. As mentioned in~\cite{zheng2021alleviating}, the high sensitivity of small bronchi will lead to the over-segmentation of airway which is also not fully addressed in this work. This leads to the higher FPR compared with the best methods. Moreover, the manual annotation can not be perfect due to the existence of complicated airway trees and the limited spacing resolution. These may not be severe issues in the segmentation of large organs while the small changes of ground-truth in the fifth/sixth generations of airway can lead to large variations of local sensitivity of foreground. To overcome this problem, the topology priors should also consider the confidence of ground-truth labels in the future.

\section{Conclusion}
In this paper, we propose a Differentiable Topology-Preserved Distance Transform (DTPDT) framework for pulmonary airway segmentation. A Topology-Preserved Surrogate (TPS) learning strategy is designed to balance the topology completeness and correctness. Furthermore, we introduce the Convolutional Distance Transform (CDT) to remove the need for discrete distance operation and construct the CDT loss function that is perceptible to breakage phenomenon. Experimental results on two public datasets collaborate the proposed method succeeds in preserving the topological structures with competitive segmentation accuracy meanwhile.

\newpage
\section*{Appendix}
\subsection{Proof of Lemma 1}
\noindent The log-sum-exponential (LSE) is as follows: \\
\centerline{$\mathit{LSE}(\mathrm{d}_{1},...,\mathrm{d}_{n}) = \mathrm{log}(\mathrm{exp}(\mathrm{d}_{1}) + ... + \mathrm{exp}(\mathrm{d}_{n}))$}\\
Under the assumption of \{$\mathrm{d}_{1} = \mathrm{d}_{2} = ... = \mathrm{d}_{k} < \mathrm{d}_{k+1} \leqslant ... \,\mathrm{d}_{n}$\},
the minimum function of $\beta$-LSE is defined by:
\begin{equation*}
\resizebox{0.9\linewidth}{!}{$
\begin{aligned}
    \min\{\mathrm{d}_{1}, ... \mathrm{d}_{n}\} &= \lim\limits_{\beta \to -\infty}\mathit{LSE}_{\beta}(\mathrm{d}_{1},...,\mathrm{d}_{n})\nonumber\\ 
    &= \lim\limits_{\beta \to -\infty}\frac{1}{\beta}\mathrm{log}(\mathrm{exp}(\beta\mathrm{d}_{1}) + ... + \mathrm{exp}(\beta\mathrm{d}_{n}))\nonumber
\end{aligned}
$} 
\end{equation*}
\begin{proof}
\\
\resizebox{0.9\linewidth}{!}{$
    \begin{aligned}
        &\lim\limits_{\beta \to -\infty} \mathit{LSE}_{\beta}(\mathrm{d}_{1},...,\mathrm{d}_{n})\nonumber\\
        &= \lim\limits_{\beta \to -\infty} [\frac{1}{\beta}\mathrm{log}(k\,\mathrm{exp}(\beta\mathrm{d}_{1}) + \sum_{j = k+1}^{n}\mathrm{exp}(\beta\mathrm{d}_{j}))]\nonumber\\ 
        &= \lim\limits_{\beta \to -\infty} [\frac{1}{\beta}\mathrm{log}(k\,\mathrm{exp}(\beta\mathrm{d}_{1})) + \frac{1}{\beta}\mathrm{log}(1 + \frac{\sum_{j = k+1}^{n}\mathrm{exp}(\beta\mathrm{d}_{j})}{k\,\mathrm{exp}(\beta\mathrm{d}_{1})})]\nonumber\\
        &= \lim\limits_{\beta \to -\infty} [\frac{1}{\beta}\mathrm{log}(k) + \mathrm{d}_{1} + \frac{1}{\beta}\mathrm{log}(1 + \frac{1}{k}\sum_{j = k+1}^{n}\mathrm{exp}(\beta(\mathrm{d}_{j}-\mathrm{d}_{1})))]\nonumber\\
        &= \mathrm{d}_{1}\nonumber
    \end{aligned} 
    $}
\end{proof}
It can be derived that the limitation converges to $\mathrm{d}_{1}$, hence the \textit{Lemma 1} is proved.

\bibliographystyle{IEEEtran}
\bibliography{paper}

\end{document}